\documentclass[conference]{IEEEtran}
\usepackage[latin9]{inputenc}
\usepackage{amsmath}
\usepackage{graphicx}

\makeatletter

\def\ps@IEEEtitlepagestyle{
  \def\@oddfoot{\mycopyrightnotice}
  \def\@evenfoot{}
}
\def\mycopyrightnotice{
  {\footnotesize 978-1-6654-3396-9/21/\$31.00~\copyright~2021 IEEE\hfill} % <--- Change here
  \gdef\mycopyrightnotice{}
}

\@ifundefined{showcaptionsetup}{}{
 \PassOptionsToPackage{caption=false}{subfig}}
\usepackage{subfig}
\makeatother

\usepackage{eso-pic}

\usepackage{avss}
\usepackage{times}
\usepackage{epsfig}
\usepackage{graphicx}
\usepackage{amsmath}
\usepackage{amssymb}
\usepackage{array}
\newcolumntype{P}[1]{>{\centering\arraybackslash}p{#1}}

\usepackage[pagebackref=true,breaklinks=true,letterpaper=true,colorlinks,bookmarks=false]{hyperref}

\avssfinalcopy % *** Uncomment this line for the final submission

\def\avssPaperID{68} % *** Enter the AVSS Paper ID here

\ifavssfinal\fi
\begin{document}

%%%%%%%%% TITLE
\title{ARPD: Anchor-free Rotation-aware People Detection using Topview Fisheye Camera}

\author{Quan Nguyen Minh $^{1,2}$, Bang Le Van $^2$,  Can Nguyen $^2$, Anh Le $^3$ and Viet Dung Nguyen $^1$\\
{\tt\small $^{1}$ Hanoi University of Science and Technology}\\
{\tt\small dung.nguyenviet1@hust.edu.vn}\\
{\tt\small $^{2}$ Viettel High Technology Industries Corporation}\\
{\tt\small {\{quannm23,banglv1,cannn1\}@viettel.com.vn}}\\
{\tt\small $^{3}$ FPT University}\\
{\tt\small anhldhe130082@fpt.edu.vn}
}
\maketitle

% \thispagestyle{empty}
%%%%%%%%% ABSTRACT
\begin{abstract}
    People detection in top-view, fish-eye images is challenging as people in fish-eye images often appear in arbitrary directions and are distorted differently. Due to this unique radial geometry, axis-aligned people detectors often work poorly on fish-eye frames. Recent works account for this variability by modifying existing anchor-based detectors or relying on complex pre/post-processing. Anchor-based methods spread a set of pre-defined bounding boxes on the input image, most of which are invalid. In addition to being inefficient, this approach could lead to a significant imbalance between the positive and negative anchor boxes. In this work, we propose ARPD, a single-stage anchor-free fully convolutional network to detect arbitrarily rotated people in fish-eye images. Our network uses keypoint estimation to find the center point of each object and regress the object's other properties directly. To capture the various orientation of people in fish-eye cameras, in addition to the center and size, ARPD also predicts the angle of each bounding box. We also propose a periodic loss function that accounts for angle periodicity and relieves the difficulty of learning small-angle oscillations. Experimental results show that our method competes favorably with state-of-the-art algorithms while running significantly faster. 
\end{abstract}
%%%%%%%%% BODY TEXT
\section{Introduction}
\begin{figure}[t]
\begin{center}
\includegraphics[width=0.95\linewidth]{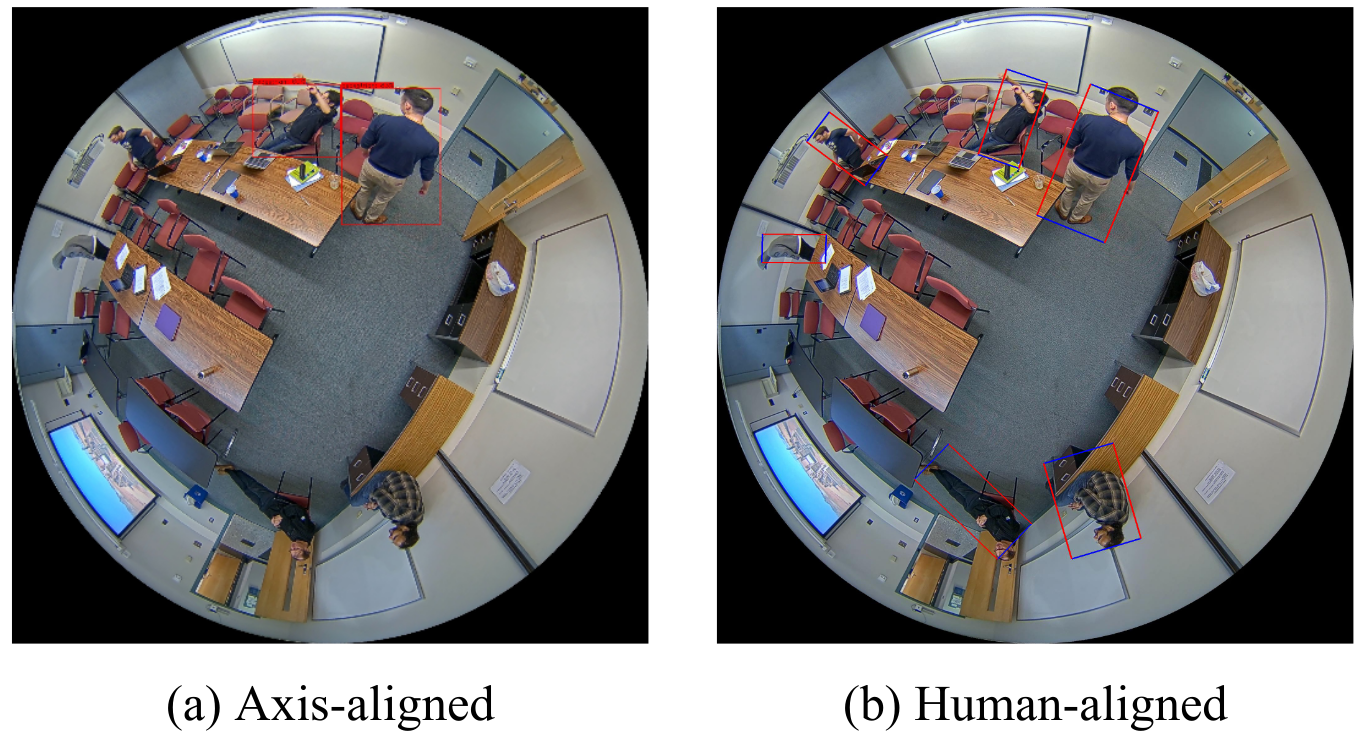}
\end{center}
   \caption{Illustration of typical people-detection results on ceiling-mounted, fish-eye images of conventional person detector (a) and the proposed method (b). Human-aligned bounding boxes fit bodies more accurately compared to axis-aligned bounding boxes.}
   \label{fig:compare}
\end{figure}

 Omni-directional cameras, most notably fish-eye cameras, are widely used for surveillance applications. One top-view fish-eye camera covers the same area as many conventional perspective cameras due to their 360$^{\circ}$ field of view. Another advantage of a ceiling-mounted fish-eye camera is minimal occlusion between objects in a frame. However, images from overhead-mounted fish-eye cameras may have human bodies at various orientations and poses. Another difficulty is severe geometric distortions at the peripheral areas of a fish-eye image, which can cause an object$\text{'}$s appearance to varying. Pedestrian detectors trained with perspective images \cite{redmon2016you,liu2016ssd,ren2015faster} cannot accommodate these variations and deformations, hence do not work well on top-view fish-eye images, often missing non-up-right bodies. (Fig. \ref{fig:compare}a)\\   
%------------------------------------------------------------------------- 

\begin{figure}[t]
\begin{center}
\includegraphics[width=1\linewidth]{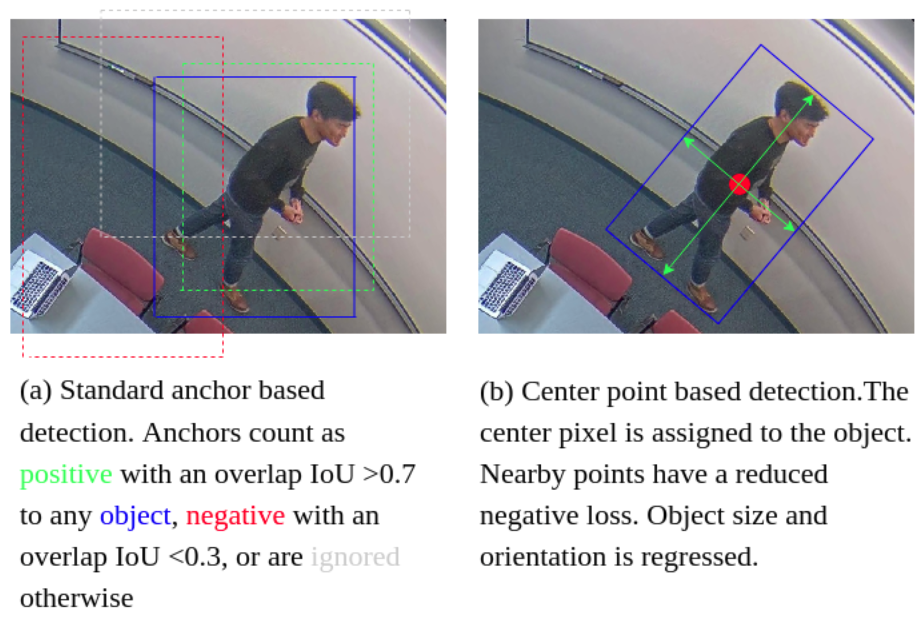}
\end{center}
   \caption{Difference between anchor-based detectors (a) and our center point detector (b). Best viewed in color. As horizontal proposals are located along the image edge, the extracted feature of an object may contain features of background and nearby objects.}
   \label{fig:anchorvskey}
\label{fig:long}
\label{fig:onecol}
\end{figure}

\indent Various methods have been proposed to tackle these problems. In early works, features are extracted using rudimentary methods such as background subtraction or edge detection \cite{choi2015real,bas2008towards}. These features have poor discrimination power and are sensitive to environmental changes and noise. Another approach utilizes traditional person detectors such as HOG and LBP, making slight adjustments to accommodate fish-eye geometry \cite{dollar2014fast,chiang2014human,krams2017people}. Recently, CNN-based people detection methods have been proposed for over-head, fish-eye images \cite{li2019supervised,tamura2019omnidirectional,duan2020rapid}. To deal with the radial geometry, these works often require complex pre-/post processing. For example, Li \etal \cite{li2019supervised} applied YOLOv3 \cite{redmon2016you} to 24 rotated, overlapping windows, and the results are re-mapped to the original image. However, this requires YOLOv3 to run 24 times for each frame. Tamura \etal \cite{tamura2019omnidirectional} tries to train a rotation-invariant model by introducing rotation augmentation during training. RAPiD \cite{duan2020rapid} modifies YOLOv3 to detect people in fish-eye images using oriented bounding boxes. However, applying horizontal-anchor-based detectors such as YOLOv3 or Faster-RCNN \cite{ren2015faster} to the oriented object detection task would lead to misalignment between the extracted features and object$\text{'}$s features (Fig. \ref{fig:anchorvskey}a).  A straightforward way to deal with this problem is to use oriented anchor boxes \cite{yang2018automatic,ding2018learning,yang2019r3det}. However, rotated anchors need to take the predefined angle (or orientation of object) into account. Hence, the mumber of anchors can be dramatically increased, increasing computational cost. Furthermore, the vast imbalance between the positive and negative anchors would lead to slow training and inferior performance. \cite{duan2019centernet}\\  

\indent In this paper, we propose ARPD, a novel single-stage anchor-free convolutional neural network that detects arbitrarily rotated bounding boxes of people in top-view fish-eye images. Our work extends the model proposed in CenterNet \cite{zhou2019objects}, one of the novel keypoint-based object detection algorithms \cite{law2018cornernet,zhou2019bottom,zhou2019objects} for standard images. We model each object as the center point of its bounding box. Other properties are directly inferred from the center keypoint feature. To capture the various alignment of people in ceiling-mounted fish-eye images, we predict the angle of each bounding box in addition to center and size. We also introduce a new rotation-aware periodic loss function that takes into account angle periodicity. The addition of an orientation head and a novel angle loss function allows ARPD to directly infer the oriented bounding boxes in fish-eye images without the need for prior boxes or non-maximal suppression. We evaluate the performance of ARPD on three publicly available person detection datasets captured by ceiling-mounted fish-eye cameras: HABBOF \cite{li2019supervised}, Mirror World \cite{ma2019mirror} and CEPDOF \cite{duan2020rapid}. The main contributions of this paper can be summarized as follows:
\begin{itemize}
	\item We propose ARPD, a single-stage anchor-free detector for rotation-aware people detection in overhead fish-eye images. Our method eliminates the need for multiple anchors and complex pre/post-processing. In experiments on multiple fish-eye datasets, ARPD achieved competitive performance compared to state-of-the-art methods and keeps a real-time inference speed.
	\item We introduce a periodic loss function that allows our network to learn the symmetry of rotated people in top-view fish-eye images. Our loss function does not suffer from the training instability and performance degeneration caused by the loss discontinuity when using standard regression functions. 
\end{itemize}

\begin{figure*}
\begin{center}
\includegraphics[width=0.95\linewidth]{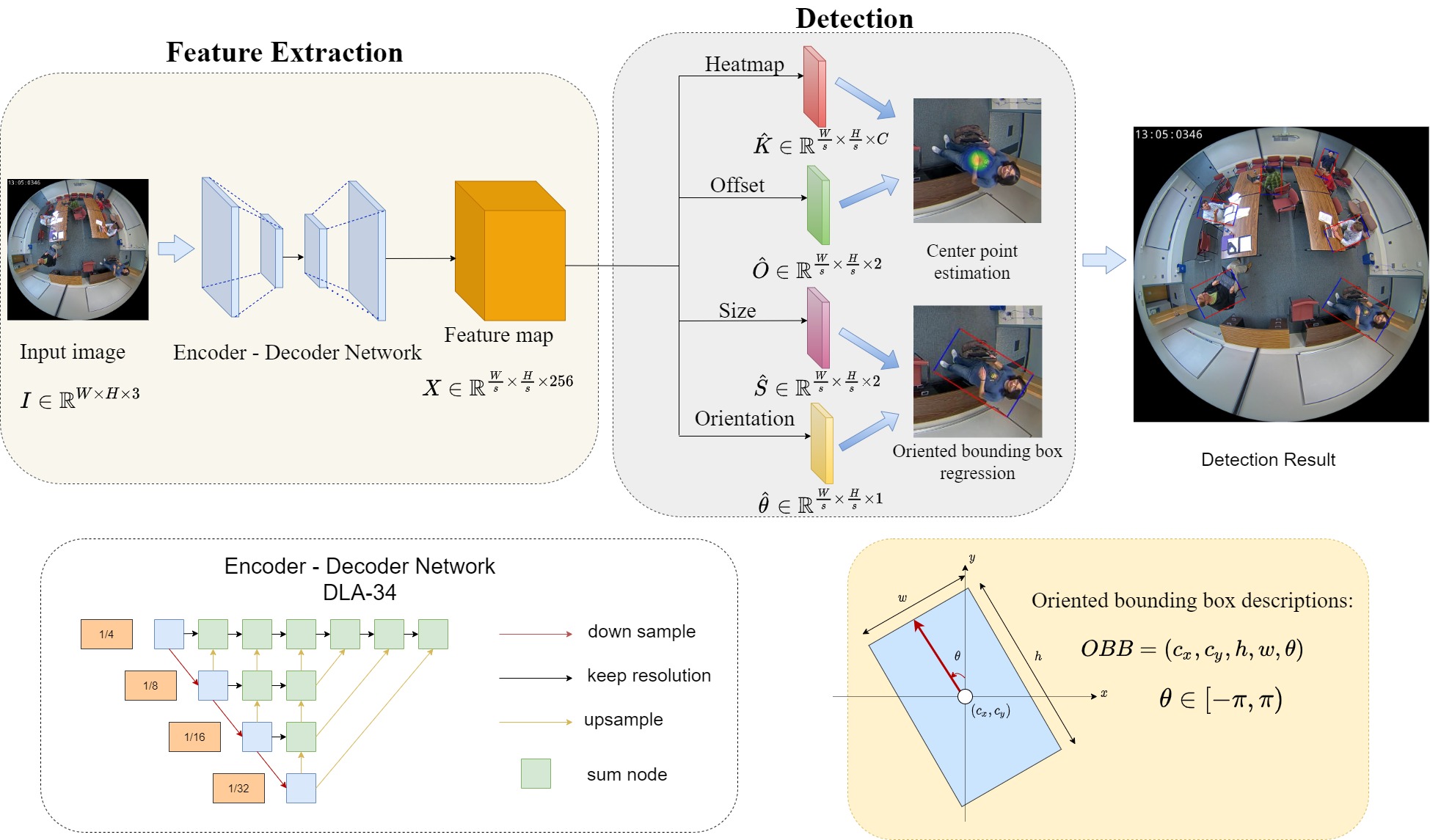}
\end{center}
   \caption{The overall architecture and the oriented bounding box (OBB) descriptions of the proposed method. We use a modified version of DLA-34 \cite{yu2018deep} as the encoder-decoder network as it gives the best speed and accuracy tradeoff \cite{zhou2019objects}. We replace the upsampling layers with 3x3 deformable convolution layers and add more skip connections from the lower layers to help increase the feature map resolution symmetrically. The output feature map is transformed into four branches: heatmap, off-set, size, and orientation. In addition to center point and size, the OBB is also represented by its angle of rotation.}
   \label{fig:architecture}
\label{fig:short}
\end{figure*}
\section{Related Works}
\label{section:RelatedWorks}
\noindent\textbf {Horizontal object detection:} Horizontal object detectors can be roughly classified into two categories: anchor-based and keypoint-based. The former consists of two-stage and one-stage methods. Two-stage detectors, most notably models from the R-CNN family \cite{girshick2015fast,ren2015faster,girshick2014rich} are region-based. First, the model proposes a set of regions of interest. Then a classifier only processes the region candidates. One-stage detectors such as YOLOv3 \cite{redmon2016you} or SSD \cite{liu2016ssd} skips the region proposal stage and runs detection directly over a dense sampling of possible locations, known as anchor boxes. Recently, keypoint-based detectors have been proposed to overcome the disadvantages of anchor-based solutions. CornerNet \cite{law2018cornernet} predicts the upper-left and lower-right corners of bounding boxes for every pixel along with an embedding, which is then used to determine the objects. ExtremeNet \cite{zhou2019bottom} predicts the center of objects as well as farthest left, right, top, and bottom points. These points are then matched based on their geometry. However, the post-grouping process is time-consuming. Zhou's CenterNet \cite{zhou2019objects} considers the center of a box as an object as well as a key point and then uses this predicted center to find the coordinates/offsets of the bounding box without post-grouping, thus making prediction faster. The keypoint-based object detectors show advantages over the anchor-based ones in terms of speed and accuracy. \cite{zhou2019objects} \\

\noindent\textbf {Oriented Object Detection:} Different from horizontal object detectors, these algorithms use rotated bounding boxes to represent oriented objects. R-DFPN \cite{yang2018automatic} designs a rotation anchor strategy to predict the minimum circumscribed rectangle of the object and build dense connections to create high-level semantic feature maps for all scales. RoI Transformer \cite{ding2018learning} designed a Rotated RoI learner to transform a Horizontal Region of Interest into a Rotated Region of Interest. To solve the misalignment problems between the region of interest and the object's feature,  R3Det \cite{yang2019r3det} proposes an end-to-end refined single-stage rotated object detector. All of the methods above use five parameter coordinates to describe oriented bounding box: center, width, height and rotation angle, with the angle defined in $\left[-\frac{\pi}{2},0\right]$ and use common regression loss. However, due to the inherent symmetry of rotated bounding boxes, this approach will lead to loss discontinuity (i.e., loss value will jump when the angle reaches its range boundary) and instability during training.\\

\noindent\textbf{People detection using overhead, fish-eye cameras:} Person detection methods using ceiling-mounted fish-eye cameras have been much less studied than conventional algorithms using standard perspective cameras, with most research appearing in recent years. Early works use techniques such as background subtraction or optical flow to obtain the location of moving people \cite{bas2008towards,choi2015real}. Methods based on Histogram of Gradients (HOG) or Local Binary Patterns (LBP) \cite{dollar2014fast,krams2017people,chiang2014human} have also been put forward. Chiang and Wang \cite{chiang2014human} applied HOG descriptor with SVM classifier on sections sliding windows extracted from fish-eye frames. In a recent work, instead of directly de-warping the fish-eye images, Krams \etal \cite{krams2017people} de-warp features extracted from the fish-eye image using an ACF classifier.\\

\indent Recently, deep learning-based algorithms have been applied to person detection in fish-eye images \cite{li2019supervised,duan2020rapid,tamura2019omnidirectional}. Li \etal \cite{li2019supervised} proposes a method in which YOLOv3 is applied to 24 rotated, overlapping windows, and the results are post-processed to produce detection results. One of the top-performing algorithms, RAPiD \cite{duan2020rapid} propose a fully convolutional network also based on YOLOv3 that detects people in fish-eye images using rotated bounding boxes. All of the methods mentioned above either require excessive computation or make modifications to horizontal anchor-based detectors, which are prone to feature misalignment problems as well as severe imbalance issues. 
\section{Proposed Method}
\label{section:ProposedMethod} 
We propose ARPD, a novel single-stage anchor-free CNN that extends Zhou$\text{'}$s CenterNet \cite{zhou2019objects}. Unlike Centernet which only predicts the location and size of each object, ARPD also predicts the angle of bounding boxes of people in a top-down, fish-eye image. We also introduce a periodic loss function based on an extension of common smooth L1 loss to deal with the loss discontinuity caused by angle periodicity.\\ 
In this section, we first describe the overall architecture
of the proposed method and the output maps in detail. The
results of the output maps are then gathered and decoded
to determine the center-point, size, and orientation of each
person in an overhead fish-eye image.\\

\subsection{ Network Architecture}
\indent The proposed network (see Fig. \ref{fig:architecture}) consists of a feature extraction network and a bounding box regression network , also known as detection head. The feature extraction network takes an image $I \in \mathbb{R}^{W \times H \times 3}$ as input and returns an output feature map. The output feature map $ X \in \mathbb{R}^{\frac{W}{s} \times \frac{H}{s} \times{256}}$ is then transformed into four branches: heatmap  ($\hat{K} \in \mathbb{R}^{\frac{W}{s} \times \frac{H}{s} \times C}$), off-set ($\hat{O} \in \mathbb{R}^{\frac{W}{s} \times \frac{H}{s} \times 2}$), size ($\hat{S} \in \mathbb{R}^{\frac{W}{s} \times \frac{H}{s} \times 2}$) and orientation ($ \hat{\theta} \in \mathbb{R}^{\frac{W}{s} \times \frac{H}{s} \times 1}$), where $C$ is the number of classes and $s = 4$ refers to the stride. For the person detection task, $C=1$. The output maps are gathered and decoded to generate the oriented bounding boxes of people in overhead fish-eye images.

\subsection{ Center point estimation }

Given an input image $I \in \mathbb{R}^{W \times H \times 3}$, we first create a keypoint heatmap $\hat{K} \in \mathcal[0,1] ^ {\frac{W}{s} \times \frac{H}{s} \times C}$. Here $\hat{K}$ is the function of $x, y, c$. A prediction $\hat{K}_{x, y, c}=1$ corresponds to detected center for class $c$. $\hat{K}_{x, y, c} = 0$ is considered as background.\\
\indent To generate ground-truth heatmaps for training, each key-point ground-truth $p \in \mathbb{R}^{2} $ are converted to low-resolution equivalent $\tilde{p}=\left\lfloor\frac{p}{s}\right\rfloor$. These centers are then splat using a $2D$ Gaussian Kernel $\exp \left(-\frac{\left(p_{x}-\tilde{p}_{x}\right)^{2}+\left(p_{y}-\tilde{p}_{y}\right)^{2}}{2 \sigma_{p}^{2}}\right)$, where $\sigma$ is an object-size adaptive standard deviation \cite{law2018cornernet}. The final result is the ground truth heatmap ${K} \in \mathcal[0,1] ^ {\frac{W}{s} \times \frac{H}{s} \times C}$. In case two or more Gaussians of the same class overlap, we get the element-wise maximum \cite{cao2019openpose}.\\
\indent When training the keypoint estimator, directly learning the positive center points would be difficult due to the imbalance between the positive and negative samples. To handle this problem, we decrease the penalty for the points inside the Gaussian bumps and use focal loss \cite{lin2017focal} to train the heatmap:
\begin{figure}[t]
\begin{center}
\includegraphics[width=0.7\linewidth]{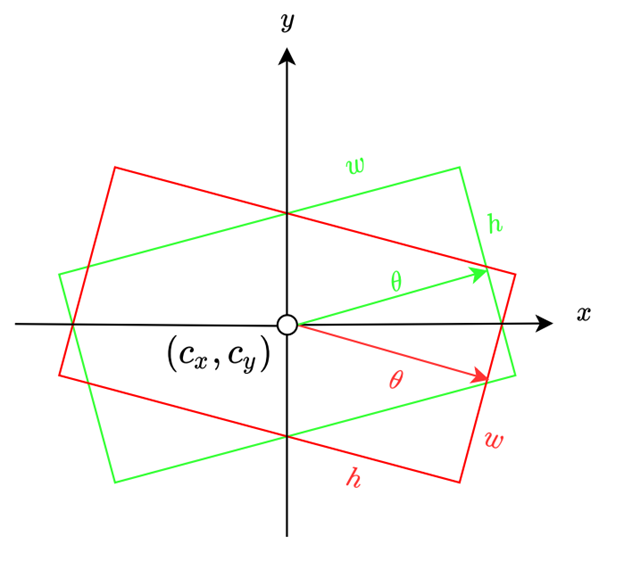}
\end{center}
    \caption{Demonstration of the loss discontinuity. Given the ground truth bounding box $\textcolor{red}{b_{gt}}=(c_{x},c_{y},h,w,-1^{\circ})$ and the predicted bounding box $\textcolor{green}{b_{pred}}=(c_{x},c_{y},w,h,89^{\circ})$. Due to the way rotated bounding box is represented traditionally, cost value is large even though prediction is close to ground truth.}
    \label{fig:discontinuity}
\label{fig:long}
\label{fig:onecol}
\end{figure}
\begin{equation}
\mathcal{L}_{K}=\frac{-1}{N} \sum_{x y c}\left\{\begin{array}{cl}
\left(1-\hat{K}_{x y c}\right)^{\alpha} \log \left(\hat{K}_{x y c}\right) & \text { if } K_{x y c}=1 \\
\left(1-K_{x y c}\right)^{\beta}\left(\hat{K}_{x y c}\right)^{\alpha} & \\
\log \left(1-\hat{K}_{x y c}\right) & \text { otherwise }
\end{array}\right.
\end{equation}
 where N is the number of objects in image I. $\alpha$ and $\beta$ are hyper-parameters of the focal loss. We use $\alpha$ = 2 and $\beta$ = 4 empirically as in \cite{law2018cornernet} in all our experiments.\\ 
\indent After the extraction of peak points from heatmaps, we have to map these coordinates to a higher dimensional input image. This will cause a discretization error as the original image pixel indices are integers and we will be predicting the float values. The offset can be calculated as $\mathbf{\hat{o}}=\left(\frac{\bar{p}_{x}}{s}-\left\lfloor\frac{\bar{p}_{x}}{s}\right\rfloor, \frac{\bar{p}_{y}}{s}-\left\lfloor\frac{\bar{p}_{y}}{s}\right\rfloor\right)$. To predict this value, we use an offset predictor $\hat{O} \in \mathbb{R}^{\frac{W}{s} \times \frac{H}{s} \times 2}$ . This offset predictor is optimized with L1 loss: 
\begin{equation}
\mathcal{L}_{off}=\frac{1}{N} \sum_{k=1}^{N} 
|{\mathbf{o}_{k}-\hat{\mathbf{o}}_{k}}|
\end{equation}
\indent In the next section, we will show how to extend this key-point estimator to the arbitrarily oriented person detection task. 
\subsection{ Oriented bounding box regression}
\label{section:OBB}
Let $\left(x_{1}^{k}, y_{1}^{k}, x_{2}^{k}, y_{2}^{k}\right)$ be the bounding box of object $k$. In addition to predicting the center point, we directly regress size $\hat{s}_{k}=\left(x_{2}^{k}-x_{1}^{k}, y_{2}^{k}-y_{1}^{k}\right)$ of each object. For this, we train a dimension head $\hat{S} \in \mathbb{R}^{\frac{W}{s} \times \frac{H}{s} \times 2}$ using standard L1 distance norm: 
\begin{equation}
\mathcal{L}_{\text {size }}=\frac{1}{N} \sum_{k=1}^{N}\left|\hat{s}_{k}-s_{k}\right|
\end{equation} 

\indent To capture the oriented bounding boxes, we also need to predict angle of an bounding box from it's center point. We define the orientation map as $\hat{\theta} \in \mathbb{R}^{\frac{H}{s} \times \frac{W}{s} \times {1}}$. To train this orientation map , we first define the ground-truth of the orientation class. Previous works on oriented object detection often use a 5-component vector $(c_{x},c_{y},w,h,\theta)$ to represent the ground truth of rotated bounding boxes, where $\theta \in \left[-\frac{\pi}{2},0\right]$. Due to rotational symmetry, bounding box $b_{1}=(c_{x},c_{y},w,h,\theta)$ with center points $c_{x},c_{y}$, width $w$, height $h$ and angle $\theta$ is indistinguishable from bounding box $b_{2}=(c_{x},c_{y},h,w,\theta - \frac{\pi}{2})$ with width $h$, height $w$ and angle $\theta- \frac{\pi}{2}$. However, common regression loss do not account for this symmetry, hence can lead to large cost even when prediction is close to ground truth (Fig. \ref{fig:discontinuity}). We solve this by enforcing the ground truth angle $\theta$ to be in range $[-\frac{\pi}{2},\frac{\pi}{2})$, and that $w$ is always smaller than $h$. Given $\widehat{t_{\theta}}$ as the output of the orientation map, we can compute the angle prediction as follow:
\begin{equation}
{\hat{\theta}}=\pi*\operatorname{Tanh}\left(\widehat{t_{\theta}}\right)
\end{equation}

\indent We limit the predicted angle $\hat{\theta}$ to be in range $[-\pi,\pi)$. This parametrization will later be explained in Section \ref{section:lossfunction}.\\
\begin{figure}[t]
\begin{center}
\includegraphics[width=1\linewidth]{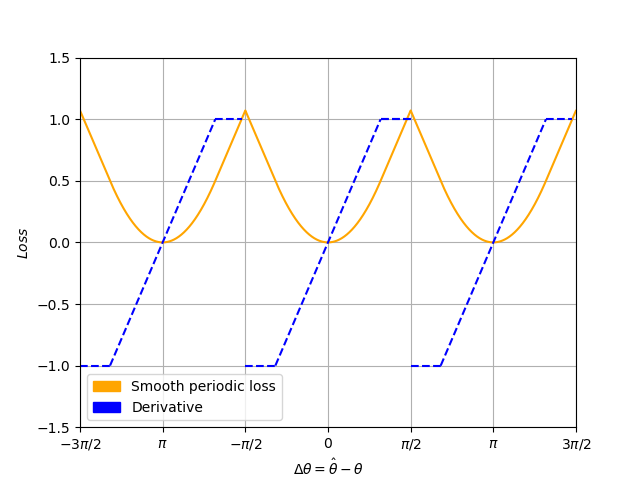}
\end{center}
    \caption{Periodic loss function with smooth L1 norm and its derivative. }
    \label{fig:lossderiv}
\label{fig:long}
\label{fig:onecol}
\end{figure}

\subsection{ Rotation-Aware Loss Function}
\label{section:lossfunction}
Our loss function is inspired by that used in CenterNet\cite{zhou2019objects}, with an additional periodic loss for angle prediction: 
\begin{equation}
\mathcal{L}_{d e t}=\mathcal{L}_{K}+\lambda_{\text {size }} \mathcal{L}_{\text {size }}+\lambda_{\text {off }} \mathcal{L}_{\text {off }}+\lambda_{\text {angle }} \mathcal{L}_{\text {angle }}
\end{equation}
where ${\mathcal{L}_{angle}}$ is the angle loss function and $\lambda_{\text {size }},\lambda_{\text {off }},\lambda_{\text{angle}}$ are constants. Traditionally, common regression functions based on L1 or L2 \cite{yang2018automatic,ding2018learning,yang2019r3det} are used for angle prediction. However, these loss functions are not take into account the fact that since a bounding box remains identical after a rotation of $ {k}*\pi$, the angle loss function must also satisfy that ${{\mathcal{L}_{angle}}\Delta{\theta}} ={{\mathcal{L}_{angle}}(\Delta{\theta}+k\pi})$, where $\Delta{\theta}$ is the difference between the ground truth and the predicted angle.\\
\\
\indent To this end, we propose a new, periodic angle loss function:
\begin{equation}
\mathcal{L}_{angle}=\frac{1}{N} \sum_{k=1}^{N}\left\{\begin{array}{cl}
|\Delta \theta_{\text {periodic }}| -0.5  & \text { if } |\Delta \theta_{\text {periodic }}| \geq1 \\
0.5*{\Delta \theta_{\text {periodic }}}^{2} & \text { otherwise }
\end{array}\right.
\end{equation}
where
\begin{equation}
\Delta \theta_{\text {periodic }}=\arctan \left(\frac{\sin \Delta \theta}{\cos \Delta \theta}\right)
\end{equation}

\begin{figure}[t]
\begin{center}
\includegraphics[width=0.8\linewidth]{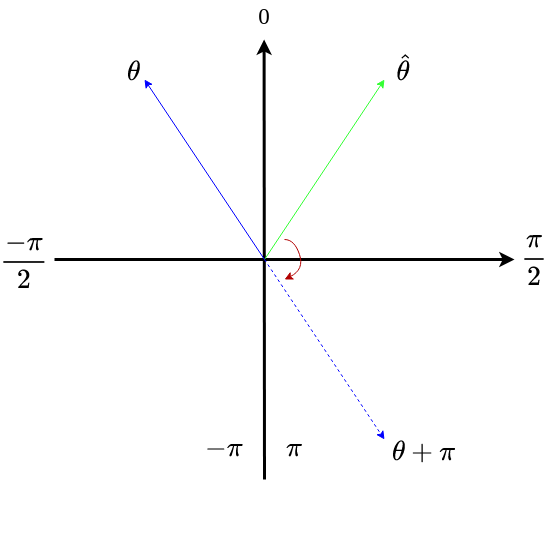}
\end{center}
    \caption{In certain cases, gradient descent will cause the predicted angle $\hat{\theta}$ (green arrow) to move further away from the ground truth angle ${\theta}$ (blue arrow). In this case, we want the network to learn to predict $\theta + \pi$ instead of $\theta$. To facilitate this behavior, we need to extend the angle range to include $\theta + \pi$ (dashed blue arrow) otherwise $\hat{\theta}$ will stop at $\frac{\pi}{2}$. }\label{fig:gradientdescent}
\label{fig:long}
\label{fig:onecol}
\end{figure}
\begin{table*}[h!]
\centering
    \caption{Performance comparision of ARPD with other state of the art methods on 3 fish-eye datasets\ref{tab:datasets}. P, R and F1 denote Precision, Recall and F1-score, respectively. Results and inference speed of ARPD and RAPiD is measured without any test time augmentation at confidence threshold $t_{conf} = 0.3$ using a single GTX 1070 Ti GPU. Results and FPS$^*$ of Tamura \etal \cite{tamura2019omnidirectional} and Li \etal \cite{li2019supervised} are as tested in RAPiD\cite{duan2020rapid}.\\}
    \label{tab:benchmark}
\small
\begin{tabular}{llllllllllllll}
\cline{3-14}
                             & \multicolumn{1}{l|}{}     & \multicolumn{4}{l|}{~~~~~~~~~~~~~~~~~~~~~~MW}                                                & \multicolumn{4}{l|}{~~~~~~~~~~~~~~~~~HABBOF}                                            & \multicolumn{4}{l|}{~~~~~~~~~~~~~~~~~CEPDOF}                                            \\ \hline
\multicolumn{1}{|l|}{}       & \multicolumn{1}{l|}{FPS}  & \multicolumn{1}{l|}{AP$_{50}$} & P     & R     & \multicolumn{1}{l|}{F1}    & \multicolumn{1}{l|}{AP$_{50}$} & P     & R     & \multicolumn{1}{l|}{F1}    & \multicolumn{1}{l|}{AP$_{50}$} & P     & R     & \multicolumn{1}{l|}{F1}    \\ \hline
\multicolumn{1}{|l|}{Tamura \etal \cite{tamura2019omnidirectional}} & \multicolumn{1}{l|}{6.8$^*$}  & \multicolumn{1}{l|}{78.2} & 0.863 & 0.759 & \multicolumn{1}{l|}{0.807} & \multicolumn{1}{l|}{87.3} & 0.970 & 0.827 & \multicolumn{1}{l|}{0.892} & \multicolumn{1}{l|}{61.0} & 0.884 & 0.526 & \multicolumn{1}{l|}{0.634} \\
\multicolumn{1}{|l|}{Li \etal . AA \cite{li2019supervised}}     & \multicolumn{1}{l|}{0.3$^*$}  & \multicolumn{1}{l|}{88.4} & 0.939 & 0.819 & \multicolumn{1}{l|}{0.874} & \multicolumn{1}{l|}{87.7} & 0.922 & 0.867 & \multicolumn{1}{l|}{0.892} & \multicolumn{1}{l|}{73.9} & 0.896 & 0.638 & \multicolumn{1}{l|}{0.683} \\
\multicolumn{1}{|l|}{Li \etal . AB \cite{li2019supervised}}     & \multicolumn{1}{l|}{0.2$^*$}  & \multicolumn{1}{l|}{95.6} & 0.895 & 0.902 & \multicolumn{1}{l|}{0.898} & \multicolumn{1}{l|}{93.7} & 0.881 & 0.935 & \multicolumn{1}{l|}{0.907} & \multicolumn{1}{l|}{76.9} & 0.884 & 0.694 & \multicolumn{1}{l|}{0.743} \\
\multicolumn{1}{|l|}{RAPiD \cite{duan2020rapid} (608)}  & \multicolumn{1}{l|}{11.9}  & \multicolumn{1}{l|}{96.6} & 0.951 & 0.931 & \multicolumn{1}{l|}{0.941} & \multicolumn{1}{l|}{97.3} & 0.984 & 0.935 & \multicolumn{1}{l|}{0.958} & \multicolumn{1}{l|}{82.4} & 0.921 & 0.719 & \multicolumn{1}{l|}{0.793} \\
\multicolumn{1}{|l|}{RAPiD \cite{duan2020rapid} (1024)}  & \multicolumn{1}{l|}{6.3}  & \multicolumn{1}{l|}{96.7} & 0.919 & 0.951 & \multicolumn{1}{l|}{0.935} & \multicolumn{1}{l|}{98.1} & 0.975 & 0.963 & \multicolumn{1}{l|}{0.969} & \multicolumn{1}{l|}{85.8} & 0.902 & 0.795 & \multicolumn{1}{l|}{0.836} \\
\multicolumn{1}{|l|}{\textbf{ARPD} (Ours)}   & \multicolumn{1}{l|}{\textcolor{red}{\textbf{21.2}}} & \multicolumn{1}{l|}{\textcolor{red}{\textbf{96.1}}} & 0.935 & 0.906 & \multicolumn{1}{l|}{\textcolor{red}{\textbf{0.92}}}  & \multicolumn{1}{l|}{\textcolor{red}{\textbf{95.6}}}& 0.968 & 0.927 & \multicolumn{1}{l|}{\textcolor{red}{\textbf{0.947}}} & \multicolumn{1}{l|}{\textcolor{red}{\textbf{79.8}}} & 0.889 & 0.702 & \multicolumn{1}{l|}{\textcolor{red}{\textbf{0.784}}} \\ \hline
\end{tabular}
\end{table*}
\indent Our angle loss function is $\pi$-periodic with respect to $\theta$. The function is also defined and differentiable except for angles such as $\Delta{\theta} = k\pi + \frac{\pi}{2}$ ( Fig. \ref{fig:lossderiv}). We ignore these angles during backpropagation. Since ground truth angle $\theta \in \left[-\frac{\pi}{2},\frac{\pi}{2}\right)$ as stated in Section \ref{section:OBB}, it is logical to force the predicted angle $\hat{\theta}$ to be in  the same range. However, this could lead to problem for gradient descent when $\Delta{\theta} \in \left(-\frac{\pi}{2},0\right]$. When $\Delta{\theta} \in \left(-\frac{\pi}{2},0\right]$, the derivative of the angle loss function $\mathcal{L}_{angle}$  will be negative, which will in turn cause gradient descent to move $\hat{\theta}$ away from $\theta$ (Fig. \ref{fig:gradientdescent}). Since a bounding box is identical after rotation of $\pi$, in this situation, it is preferable that the network learn to predict  $\hat{\theta} + \pi$ instead. To allow this behavior, we extend the range of predicted angle $\hat{\theta}$ to $[-\pi,\pi)$. To provide steady gradients for large values of $\Delta{\theta}$  and  less oscillations during updates when $\Delta{\theta}$ is small, we use smooth L1 norm for our angle loss (Fig. \ref{fig:lossderiv}).

\label{section:Experimentalresults}

\begin{figure*}[t]
\begin{center}
\includegraphics[width=1\linewidth]{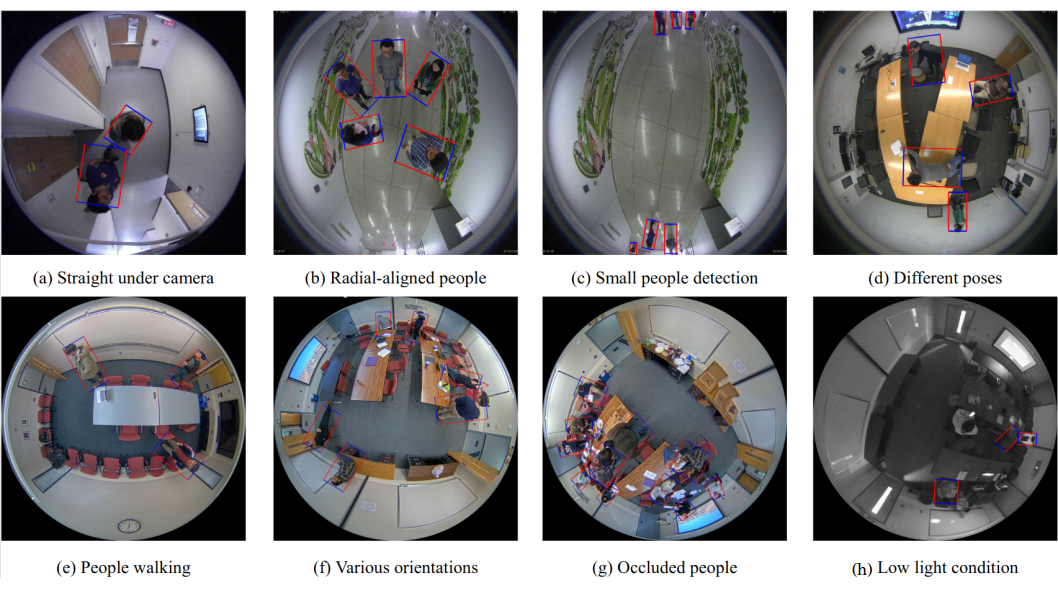}
\end{center}
    \caption{ Qualitative results of ARPD on MW (a-d), HABBOF (e) and CEPDOF (f-h). ARPD works well on both easy and challenging cases, such as heavy occlusion, various poses and background. For cases of small people detection (c), in some cases the bounding boxes does not fully enclose the person. Unsurprisingly, low-light scenarios (h) remains challenging.}
    \label{fig:qualitative}
\label{fig:long}
\label{fig:onecol}
\end{figure*}

\section{Experimental results}
\subsection{ Datasets and Implemetation details}
\indent We evaluate our method on three publicly available fish-eye datasets: Mirror World (MW) \cite{ma2019mirror}, HABBOF \cite{li2019supervised} and CEPDOF \cite{duan2020rapid}. All three datasets contain videos of people captured by ceiling-mounted fish-eye cameras in various scenarios. Further information about the datasets are given in Table \ref{tab:datasets}. To measure the performance of our algorithm, we adopt the Average Precision metric used by COCO \cite{lin2014microsoft} in addition to F1 score, precision and recall. Since a single person can be represented by multiple ground-truth rotated bounding boxes with different angles, we only consider the AP at IoU = 0.5 (AP$_{50}$). We use a confidence threshold of 0.3 for all our tests.\\

\indent For the training and testing stage, we resize the input image to $512 \times 512$. In addition to random flip, random cropping, random scaling, and color jittering, we also use
random rotation augmentation. We first train our network on MS COCO 2017 \cite{lin2014microsoft} for 90 epochs. In the training
phase, perspective images (containing people) from the COCO dataset are randomly rotated before being input into the network. We split the three datasets\ref{tab:datasets} into train/test split, i.e., two datasets are used for training, and the remaining are used for testing. For example, we fine-tune ARPD on HABBOF and CEPDOF for 10 epochs, and test it on MW. This process is repeated for all permutations. We use Adam optimizer with an initial learning rate of $1.25 \times 10^{-4}$, with learning rate drop at epochs 60 and 80. Unless otherwise specified, we set $\lambda_{\text {off}} = 1$ and $\lambda_{\text {size}} = \lambda_{\text {angle}} = 0.1$ in all of our experiments. We train the network with the batch size of 32 on two NVIDIA 2080Ti. The inference speed is measured on a single NVIDIA 1070Ti.\\
\begin{table}[!]
\centering
    \caption{Statistics of three publicly available overhead fish-eye image datasets. All images have a $360^{\circ}$ field of view and 1:1 aspect ratio.}
    \label{tab:datasets}
\begin{tabular}{|l|lll|}
\multicolumn{1}{l}{} &              &              & \multicolumn{1}{l}{}    \\ 
\hline
Dataset              & \# of videos & \# of frames & \# of GT boxes  \\ 
\hline
MW                   & ~ ~ ~ 19     & ~ ~ ~8752    & ~ ~ ~ 22758       \\
CEPDOF               & ~ ~ ~ ~4     & ~ ~ 25504    & ~ ~ ~ 173428       \\
HABBOF               & ~ ~ ~ ~8     & ~ ~ ~5873    & ~ ~ ~ 20430       \\
\hline
\end{tabular}
\end{table}
\\

\begin{table*}[h!]
\centering
    \caption{Ablation experiments conducted on Mirror World dataset. Results are shown in COCO $AP_{50}$. \\
    \\}
    \label{tab:ablation}
\begin{tabular}{|l|c|lllllll|}

\hline
\multicolumn{2}{|c|}{Method}                               & \multicolumn{7}{c|}{ARPD}                        \\ \hline
                                 & L1                      & \checkmark   &   &   &   &                          &   &   \\
\multicolumn{1}{|c|}{Angle loss} & Periodic L1             &   & \checkmark   &   &   &                          &   &   \\
                                 & Smooth periodic L1  &   &   & \checkmark   & \checkmark   & \checkmark                          & \checkmark   & \checkmark   \\ \hline
                                 & $\left[-\frac{\pi}{2},\frac{\pi}{2}\right)$                      & \checkmark   & \checkmark   & \checkmark   &   &                          &   &   \\
\multicolumn{1}{|c|}{Prediction range}                 & $(-\infty,\infty)$                        &   &   &   & \checkmark   &                          &   &   \\
                                 & $[-\pi,\pi)$                      &   &   &   &   & \checkmark                          & \checkmark   & \checkmark   \\ \hline
                                 & 1                       & \checkmark   & \checkmark   & \checkmark   & \checkmark   & \checkmark                          &   &   \\
\multicolumn{1}{|c|}{$\lambda_{\text {angle }}$}       & 0.01                    &   &   &   &   &                          & \checkmark   &   \\
                                 & 0.1                     &   &   &   &   &                          &   & \checkmark   \\ \hline
\multicolumn{2}{|c|}{$AP_{50}$}                                & 86.2  & 88.9  & 90.1  & 92.1  &              93.2            & 95.3  & \textbf{96.1}  \\ \hline
\end{tabular}
\end{table*}
\subsection{ Main results}
\indent Results from Table. \ref{tab:benchmark} shows that ARPD's performance come close to the top-performing algorithm , while running many times faster than all other methods tested. This makes ARPD superior for the real-time person detection task. Our method outperforms Tamura \etal's method \cite{tamura2019omnidirectional} by a considerable margin on all three fish-eye datasets, and is slightly better in terms of AP compared to the method of Li \etal \cite{li2019supervised}, while running tens of times faster. We achieve 
an execution speed of 21 frames per second, which is nearly two times faster than RAPiD while not having to sacrifice considerably in terms of accuracy. ARPD achieves an $AP_{50}$ score of more than 95 on MW and HABBOF, both of which mostly consist of people walking and standing appearing radially-oriented (Fig. \ref{fig:qualitative} a-e). ARPD is also capable of detecting various challenging scenarios in CEPDOF such as extreme body poses or occlusion (Fig. \ref{fig:qualitative} f-g). However, scenarios such  as low-light or small objects remain difficult (Fig. \ref{fig:qualitative} h). 
\subsection{ Ablation Experiments}
\indent In this section, we conduct various experiments to analyze how individual parts of ARPD contributes to the overall performance, and the effectiveness of novel elements we introduced. \\

\indent \textbf{Rotation aware Angle loss:} We compare the performance of our novel periodic loss function against
commonly used regression loss functions. Results from Table \ref{tab:ablation} proves that our loss function is better for oriented person detection . Smooth L1 norm also performs slightly better than L1. This can be explained due to the fact that Smooth L1 allows for better optimization when $\Delta{\theta}$ is small.\\
\\
\\
\indent \textbf{Parameterization of oriented bounding box:} As shown in Table \ref{tab:ablation}, there is a notable performance improvement when we extend the prediction range from $\left[-\frac{\pi}{2},\frac{\pi}{2}\right)$ to $[-\pi,\pi)$, however increasing it further does not significantly impact accuracy.\\

\indent \textbf{Impact of different orientation weight:} We test the approach's sensitivity to the orientation weight $\lambda_{angle}$. $\lambda_{angle} = 0.1$ yields the best result. Performance noticeably deteriorate when $\lambda_{angle} > 0.1$.
\section{Conclusions}
\label{section:Conclusions}
In this paper, we propose ARPD, a new oriented person detection method in fish-eye images based on center point detection. ARPD is single-stage, free of anchor or NMS post-processing. In addition to the location and dimension of the bounding boxes, ARPD also predicts its angle of
rotation. We also introduce an angle-aware periodic loss function based on smooth L1 norm, which considers angle periodicity and is more sensitive towards small-angle variation between ground truth and prediction. Experimental results show that our method achieves the best speed-to-accuracy trade-off on multiple fish-eye datasets. Our method would benefit real-world applications and serve as a baseline algorithm for real-time person detection on fish-eye cameras.\\


{\small
\begin{thebibliography}{10}

\bibitem{redmon2016you}
J.~Redmon, S.~Divvala, R.~Girshick, and A.~Farhadi, ``You only look once:
  Unified, real-time object detection,'' in {\em Proceedings of the IEEE
  conference on computer vision and pattern recognition}, pp.~779--788, 2016.

\bibitem{liu2016ssd}
W.~Liu, D.~Anguelov, D.~Erhan, C.~Szegedy, S.~Reed, C.-Y. Fu, and A.~C. Berg,
  ``Ssd: Single shot multibox detector,'' in {\em European conference on
  computer vision}, pp.~21--37, Springer, 2016.

\bibitem{ren2015faster}
S.~Ren, K.~He, R.~Girshick, and J.~Sun, ``Faster r-cnn: Towards real-time
  object detection with region proposal networks,'' {\em Advances in neural
  information processing systems}, vol.~28, pp.~91--99, 2015.

\bibitem{choi2015real}
Y.-W. Choi, K.-K. Kwon, J.-H. Kim, K.-J. Na, and S.-G. Lee, ``Real time
  omni-directional object detection using background subtraction of fisheye
  image,'' {\em Journal of Institute of Control, Robotics and Systems},
  vol.~21, no.~8, pp.~766--772, 2015.

\bibitem{bas2008towards}
E.~Bas, D.~Erdogmus, U.~Ozertem, and M.~Pavel, ``Towards fish-eye camera based
  in-home activity assessment,'' in {\em 2008 30th Annual International
  Conference of the IEEE Engineering in Medicine and Biology Society},
  pp.~2558--2561, IEEE, 2008.

\bibitem{dollar2014fast}
P.~Doll{\'a}r, R.~Appel, S.~Belongie, and P.~Perona, ``Fast feature pyramids
  for object detection,'' {\em IEEE transactions on pattern analysis and
  machine intelligence}, vol.~36, no.~8, pp.~1532--1545, 2014.

\bibitem{chiang2014human}
A.-T. Chiang and Y.~Wang, ``Human detection in fish-eye images using hog-based
  detectors over rotated windows,'' in {\em 2014 IEEE International Conference
  on Multimedia and Expo Workshops (ICMEW)}, pp.~1--6, IEEE, 2014.

\bibitem{krams2017people}
O.~Krams and N.~Kiryati, ``People detection in top-view fisheye imaging,'' in
  {\em 2017 14th IEEE international conference on advanced video and signal
  based surveillance (AVSS)}, pp.~1--6, IEEE, 2017.

\bibitem{li2019supervised}
S.~Li, M.~O. Tezcan, P.~Ishwar, and J.~Konrad, ``Supervised people counting
  using an overhead fisheye camera,'' in {\em 2019 16th IEEE International
  Conference on Advanced Video and Signal Based Surveillance (AVSS)}, pp.~1--8,
  IEEE, 2019.

\bibitem{tamura2019omnidirectional}
M.~Tamura, S.~Horiguchi, and T.~Murakami, ``Omnidirectional pedestrian
  detection by rotation invariant training,'' in {\em 2019 IEEE winter
  conference on applications of computer vision (WACV)}, pp.~1989--1998, IEEE,
  2019.

\bibitem{duan2020rapid}
Z.~Duan, O.~Tezcan, H.~Nakamura, P.~Ishwar, and J.~Konrad, ``Rapid:
  rotation-aware people detection in overhead fisheye images,'' in {\em
  Proceedings of the IEEE/CVF Conference on Computer Vision and Pattern
  Recognition Workshops}, pp.~636--637, 2020.

\bibitem{yang2018automatic}
X.~Yang, H.~Sun, K.~Fu, J.~Yang, X.~Sun, M.~Yan, and Z.~Guo, ``Automatic ship
  detection in remote sensing images from google earth of complex scenes based
  on multiscale rotation dense feature pyramid networks,'' {\em Remote
  Sensing}, vol.~10, no.~1, p.~132, 2018.

\bibitem{ding2018learning}
J.~Ding, N.~Xue, Y.~Long, G.-S. Xia, and Q.~Lu, ``Learning roi transformer for
  detecting oriented objects in aerial images,'' {\em arXiv preprint
  arXiv:1812.00155}, 2018.

\bibitem{yang2019r3det}
X.~Yang, Q.~Liu, J.~Yan, A.~Li, Z.~Zhang, and G.~Yu, ``R3det: Refined
  single-stage detector with feature refinement for rotating object,'' {\em
  arXiv preprint arXiv:1908.05612}, vol.~2, no.~4, 2019.

\bibitem{duan2019centernet}
K.~Duan, S.~Bai, L.~Xie, H.~Qi, Q.~Huang, and Q.~Tian, ``Centernet: Keypoint
  triplets for object detection,'' in {\em Proceedings of the IEEE/CVF
  International Conference on Computer Vision}, pp.~6569--6578, 2019.

\bibitem{zhou2019objects}
X.~Zhou, D.~Wang, and P.~Kr{\"a}henb{\"u}hl, ``Objects as points,'' {\em arXiv
  preprint arXiv:1904.07850}, 2019.

\bibitem{law2018cornernet}
H.~Law and J.~Deng, ``Cornernet: Detecting objects as paired keypoints,'' in
  {\em Proceedings of the European conference on computer vision (ECCV)},
  pp.~734--750, 2018.

\bibitem{zhou2019bottom}
X.~Zhou, J.~Zhuo, and P.~Krahenbuhl, ``Bottom-up object detection by grouping
  extreme and center points,'' in {\em Proceedings of the IEEE/CVF Conference
  on Computer Vision and Pattern Recognition}, pp.~850--859, 2019.

\bibitem{ma2019mirror}
N.~Ma, ``Mirror worlds challenge,''
\newblock Retrieved from
  http://www2.icat.vt.edu/mirrorworlds/challenge/index.html.

\bibitem{yu2018deep}
F.~Yu, D.~Wang, E.~Shelhamer, and T.~Darrell, ``Deep layer aggregation,'' in
  {\em Proceedings of the IEEE conference on computer vision and pattern
  recognition}, pp.~2403--2412, 2018.

\bibitem{girshick2015fast}
R.~Girshick, ``Fast r-cnn,'' in {\em Proceedings of the IEEE international
  conference on computer vision}, pp.~1440--1448, 2015.

\bibitem{girshick2014rich}
R.~Girshick, J.~Donahue, T.~Darrell, and J.~Malik, ``Rich feature hierarchies
  for accurate object detection and semantic segmentation,'' in {\em
  Proceedings of the IEEE conference on computer vision and pattern
  recognition}, pp.~580--587, 2014.

\bibitem{cao2019openpose}
Z.~Cao, G.~Hidalgo, T.~Simon, S.-E. Wei, and Y.~Sheikh, ``Openpose: realtime
  multi-person 2d pose estimation using part affinity fields,'' {\em IEEE
  transactions on pattern analysis and machine intelligence}, vol.~43, no.~1,
  pp.~172--186, 2019.

\bibitem{lin2017focal}
T.-Y. Lin, P.~Goyal, R.~Girshick, K.~He, and P.~Doll{\'a}r, ``Focal loss for
  dense object detection,'' in {\em Proceedings of the IEEE international
  conference on computer vision}, pp.~2980--2988, 2017.

\bibitem{lin2014microsoft}
T.-Y. Lin, M.~Maire, S.~Belongie, J.~Hays, P.~Perona, D.~Ramanan,
  P.~Doll{\'a}r, and C.~L. Zitnick, ``Microsoft coco: Common objects in
  context,'' in {\em European conference on computer vision}, pp.~740--755,
  Springer, 2014.

\end{thebibliography}
}
\end{document}